# Feature Level Fusion of Biometrics Cues: Human Identification with Doddington's Caricature

Dakshina Ranjan Kisku[1], Phalguni Gupta[2], Jamuna Kanta Sing[3]

[1] Department of Computer Science and Engineering,
Dr. B. C. Roy Engineering College, Durgapur – 713206, India
[2] Department of Computer Science and Engineering,
Indian Institute of Technology Kanpur, Kanpur – 208016, India
[3] Department of Computer Science and Engineering,
Jadavpur University, Kolkata – 700032, India
{drkisku, jksing}@ieee.org; pg@cse.iitk.ac.in

**Abstract.** This paper presents a multimodal biometric system of fingerprint and ear biometrics. Scale Invariant Feature Transform (SIFT) descriptor based feature sets extracted from fingerprint and ear are fused. The fused set is encoded by K-medoids partitioning approach with less number of feature points in the set. K-medoids partition the whole dataset into clusters to minimize the error between data points belonging to the clusters and its center. Reduced feature set is used to match between two biometric sets. Matching scores are generated using wolf-lamb user-dependent feature weighting scheme introduced by Doddington. The technique is tested to exhibit its robust performance.

## 1 Introduction

The multimodal biometric systems [1] are found to be extremely useful and exhibit robust performance over the unimodal biometric systems in terms of several constraints. The aim of any multimodal system [1] is to acquire multiple sources of information from different modalities and minimize the error prone effect of monomodal systems. The focus to multimodal systems is the fusion of various biometric modality data at the various information fusion levels [2] such as sensor, feature extraction, matching score, rank or decision levels. In [1], [9] there exist multimodal biometrics systems based on face and fingerprint, face and voice, signature and voice, face and ear. However, the existence of any system through fusion of fingerprint [4] and ear [3] biometrics at feature extraction level is not known to the authors. Since, the fingerprint biometrics is widely used and the accuracy level of fingerprint system is high as compared to other biometric traits. Again, ear biometric is robust and effective to biometric applications. Further, ears [3] have several advantages over facial features such as uniform distributions of intensity and spatial resolution, and less variability with expressions and orientation of the face [5].

Unlike face recognition [5] with changing lightning and different pose of head positions, ear shape does not change over time and ageing. Further low effect of lighting conditions and spatial distribution of pixels has made ear biometrics an emerging authentication system. Fingerprints are established themselves as widely used and efficient biometric traits for verifying individuals. Design of a reliable fingerprint verification system depends on underlying constraints such as representation of fingerprint patterns, sensing fingerprints and matching algorithms.

This paper presents a robust feature level fusion technique of fingerprint [4] and ear [3] biometrics. It uses Scale Invariant Feature Transform (SIFT) descriptor [6] to obtain features from the normalized fingerprint and ear. These features are fused to get one feature vector. To obtain the more discriminative reduced set of feature vector, PAM (Partitioning About Medoids) characterized K-medoids clustering approach [7] is applied to the concatenated feature set. Matching scores between features of database set and that of query set are obtained by K-nearest neighbor approach [6] and Euclidean distance metric [2]. The relevance of individual matchers towards more efficient and robust performance is determined by wolf and lamb factors as discussed in [8]. Both these factors can decrease the performance of any biometric system by accepting more and more imposters as false accept. This paper extends the notions of Doddington's weighting scheme [8] in the proposed feature level fusion by adaptive user weighting process. The performance of feature level fusion has been determined on a multimodal database containing fingerprint and ear images. The results show significant improvements over the individual matching performance of fingerprint and ear biometrics as well as an existing feature level fusion scheme [2] which have used SIFT as feature descriptor.

Next section introduces SIFT descriptor for feature extraction. Extraction of SIFT features from fingerprint and ear images and fusion by concatenation of extracted SIFT features is presented in Section 3. In Section 4, PAM characterized K-medoids clustering approach is applied to the concatenated feature set to handle the curse of dimensionality. A matching score generation technique using reduced features sets obtained from gallery and probe samples is also described in this section. User-dependent matcher weighting scheme using Doddington's method with adaptive approach has been applied to the proposed feature level fusion in Section 5. Results have been analyzed in Section 6 and finally, concluding remarks are made in the last section.

## 2  Description of SIFT Features

SIFT descriptor [2], [6] has been successfully used for general object detection and matching. SIFT operator can be able to detect stable and invariant feature points in images. It is invariant to image rotation, scaling, partly illumination changes, and 3D projective transform. SIFT descriptor detects feature points efficiently through a staged filtering approach that identifies stable points in the Gaussian scale-space. This is achieved by four steps: (i) selection of candidates for feature points by searching peaks in the scale- space from a difference of Gaussian (DoG) function, (ii) localization of these points by using the measurement of their stability, (iii) assignment of orientations based on local image properties, and finally, (iv) calculation of the feature descriptors which represent local shape distortions and illumination changes. These steps can determine candidate locations and a detailed fitting is performed to the nearby data for the candidate location, edge response and peak magnitude. To achieve invariance to image rotation, a consistent orientation is assigned to each feature point based on local image properties. Histogram of orientations is formed from the gradient orientation at all sample points within a circular window of a feature point. Peaks in this histogram correspond to the dominant directions of each feature point. For illumination invariance, 8 orientation planes are defined. Finally, the gradient magnitude and the orientation are smoothened by applying a Gaussian filter and then sampled over a 4×4 grid with 8 orientation planes.

Each feature point [6] contains four types of information – spatial location ($x, y$), scale ($s$), orientation ($\theta$) and keypoint descriptor ($k$). All these feature information are used. More formally, local image gradients are measured at the selected scale in the region around each keypoint. The measured gradients' information is then transformed into a vector representation that contains a vector of 128 elements for each keypoints calculated over extracted keypoints. These keypoint descriptor vectors represent local shape distortions and illumination changes.

## 3  Feature Extraction and Feature Level Fusion

### 3.1  Preprocessing and SIFT Feature Extraction

For fingerprint verification [4], three types of features are used: (i) global ridge and furrow structure forming a special pattern in the central region of the fingerprint, (ii) minutiae details associated with local ridge and furrow structure and (iii) correlation. However, minutiae based fingerprint systems [4] show higher accuracy as compared to other two types of systems. Local texture around minutiae points is more desirable and useful for good accuracy rather than the whole fingerprint image since global texture is sensitive to non-linear and non-repeatable deformation of such images.

In the proposed method, SIFT features are extracted from the whole fingerprint image. On the other hand, ear biometric [3] has been newly introduced for identity verification and it is considered as one of the most reliable and invariant biometrics characteristics. SIFT descriptor is used to detect stable invariant points for general object recognition and it does not require generally any image to be preprocessed. However, in the proposed work, few preprocessing operations are performed on ear image to obtain better accuracy. In the first step, localization of ear image is done by detecting manually two points on ear image viz. Triangular Fossa and Antitragus [9]. Localization technique proposed in [9] has been used in this paper. In the next step fingerprint and ear images are normalized having adjustable gray level distribution. To make uniform distribution of gray levels, image intensity is measured in the central area and the distribution is adjusted accordingly. This is performed using adaptive histogram equalization technique. The proposed work uses the whole ear image for SIFT features extraction by making indifference it with the fingerprint texture. The use of SIFT descriptor not only increases the number of invariant SIFT points while feature extraction, but also increases the reliability of system by accumulating large number of points. Extraction of SIFT feature points can be controlled by local minima or maxima in a Gaussian scale space. The feature numbers can also be controlled by a set of parameters such as octaves and scales.

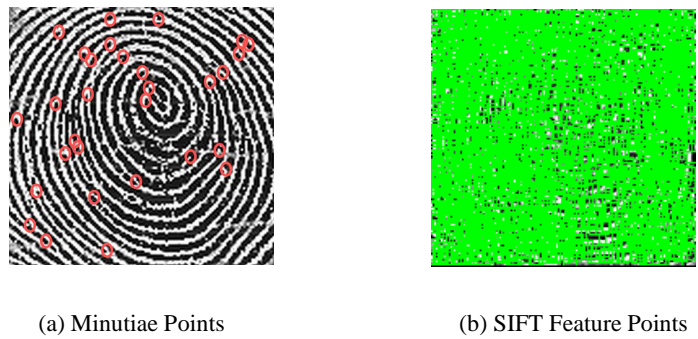

(a) Minutiae Points  (b) SIFT Feature Points

Fig. 2. Minutiae and SIFT Feature Points of a Fingerprint.

A fingerprint may contain thousand SIFT features. Figure 2 shows a typical fingerprint image from where 30 minutiae points and 2449 SIFT feature points have been detected. The number of SIFT feature points obtained from an ear may vary from hundreds to few thousands. An ear image is shown in Figure 3 from where 1154 SIFT feature points are extracted.

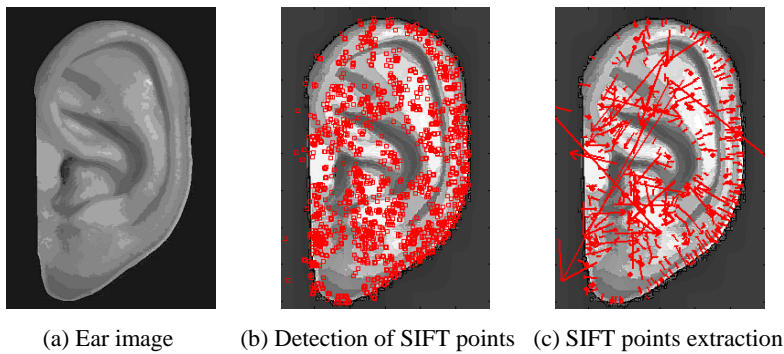

(a) Ear image  (b) Detection of SIFT points  (c) SIFT points extraction

Fig. 3. SIFT Feature Points of an Ear Image

### 3.2 Feature Level Fusion of SIFT Keypoints

Concatenation technique [2] is used to fuse SIFT features extracted from fingerprint and ear at the feature extraction level. Feature level fusion is difficult to achieve in practice because multiple modalities may have incompatible feature [1], [2] sets and the correspondence among different feature spaces may be unknown. The concatenated feature set exhibits better discrimination capability than the individual feature vectors obtained from fingerprint and ear biometrics separately.

## 4 Feature Reduction and Matching

### 4.1 Feature reduction

PAM (Partitioning About Medoids) characterized K-medoids partitioning algorithm [7] is applied to the concatenated features set to obtain the reduced set of features which can provide more discriminative and meaningful reduced set of features. This clustering algorithm is an adaptive version of K-means clustering approach. It is used to partition dataset into some groups and minimizes the squared error between the points that belong to a cluster and a point designated as the center of the cluster. K-medoids chooses data points as cluster centers (also called 'medoids'). K-medoids clusters the dataset of n objects into k clusters. It is more robust to noise and outliers as compared to K-means clustering algorithm [7]. In the proposed method, K-medoids clustering algorithm is applied to the SIFT points set, which is formed by concatenation of SIFT features extracted from fingerprint and ear images. The redundant features are removed using K-medoids clustering technique and choosing the most proximate features as the representative of the set of similar features. A medoid can be defined as the object of a cluster, which means dissimilarity to all the objects in the cluster is minimal. The most generalization of K-medoids algorithm is the Partitioning Around Medoids (PAM) algorithm which can be given below.

Step 1: *Randomly select k number of points from the concatenated SIFT points set as the medoids.*
Step 2: *Assign each SIFT feature point to the closest medoid and the closest medoid can be defined using a distance metric (Euclidean distance metric).*
Step 3: *for each medoid i, i = 1, 2...k*
 *for each non-medoid SIFT point j*
 *swap i and j and compute the total cost of the configuration*
Step 4: *Select the configuration with the lowest cost*
Step 5: *Repeat Step 1 to Step 5 until there is no change in the medoid.*

### 4.2 Matching

The optimal features are matched using the K-nearest neighbor approach [6] by computing distances from the optimal feature vector obtained from probe samples to all stored optimal features which are obtained from gallery sample and k – closest samples are selected. In the proposed experiment, by using K-NN, a set of best matched features are selected. This computation is made using spatial location (x, y), scale (s), orientation (θ) and keypoint descriptor (k) information of SIFT descriptor.

Euclidean distance is used for distance computation. The number of best matched features denotes the matching score for a particular fingerprint-ear pair sample. The matching scores are normalized in the range [0-1] [1], [8].

## 5   Adaptive Weighting using Doddington's Approach

Reliability of each fused matching score can be increased by applying the proposed adaptive Doddington's user-dependent user weighting scheme [8]. In order to decrease the number of false accepts in the proposed system, we extend the notion used for weighting the matchers by wolf-lamb concept introduced by Doddington. The authors in [8] have also used the Doddington's concept for user weighting by weights the matchers in the fused biometric system. In the proposed system, we have computed the adaptive weights by making tan-hyperbolic weight for each matcher by assigning weights to individual matching scores. The proposed adaptive weighting scheme decreases the effect of imposter users rapidly while it is compared with the method discussed in [8]. The modified Doddington's scheme is described as follows.

Let the user-dependent fused match score for user $p$ can be calculated as

$$fs_p = \sum_{ms=1}^{MS} w_p^{ms} n_p^{ms}, \forall p \qquad (1)$$

where $MS$ denotes the total number of matching scores obtained from matching of probe and gallery samples and $w_p^{ms}$ represents the weight that can be assigned to the matching score $n_p^{ms}$ for user $p$. It is assumed that the fused scores carry the wolf-lamb properties together which are not easy to determine separately. Assumptions have been made by Doddington's [8] that the users who are labeled as lambs can be imitated easily and wolves can imitate other users. Lambs and wolves – these two constraints can lead to false accepts while they degrade the performance of biometric systems. After computing weight $w_p^{ms}$ for each matcher, we extend the notions for each weight to make it adaptive one. The adaptive weight notion can be obtained by taking tan-hyperbolic of computed weights. The range of $w_p^{ms}$ weight must be [0,1] and the sum of all weights should be 1. The objective of this adaptive weighting scheme is to reduce the lambness of matchers while feature level fusion of two or more biometric traits is formulated. The adaptive weight notation has been established by extending the usual notions used by [8] and by adopting the robust statistics method [8] as follows.

$$W(w'^1_p, w'^2_p, ..., w'^{ms}_p, ..., w'^{MS}_p) = \tanh(w'^{ms}_p) \qquad (2)$$

Now the Equation (1) can be re-written using Equation (2) as

$$fs_p = \sum_{ms=1}^{MS} w'^{ms}_p \, n^{ms}_p, \forall p \qquad (3)$$

## 6   Experimental Results

The proposed technique is tested on IIT Kanpur multimodal database consisting of fingerprint and ear images acquired from 1550 subjects and each subject has provided 2 fingerprints and 2 ear images. Fingerprint images are acquired using an optical sensor at 500 dpi and the ear images are obtained using a high resolution digital camera. After normalization of fingerprint and ear images, fingerprint images are downscaled to 200×200 pixels. This high resolution to fingerprint image may increase the number of SIFT features. On the other hand, the ear images are taken under controlled environment in different sessions. The ear viewpoints are consistently kept neutral and the ear images are downscaled to 200×140 pixels. The following protocol has been established for multimodal evaluation and testing.

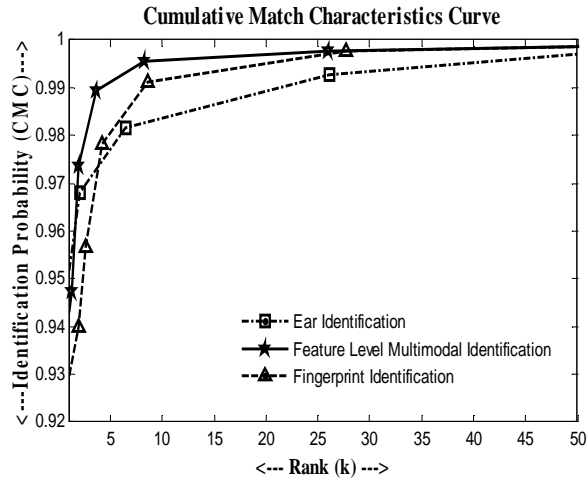

Fig. 4. Cumulative Match Characteristics Curves

**Training:** One image per person from each modality i.e., fingerprint and ear is used enrollment in gallery database and which are further used for feature extraction and feature level fusion. Fused feature vector is then encoded and is saved as gallery vector and is used for identification and verification.

**Testing:** Pair of fingerprint and ear images is used for testing. Imposter matching scores are generated by validating and testing the first client against itself and also against the remaining subjects. Fused feature vector is generated from a pair of fingerprint and ear images and is compared with the gallery feature vectors.

Rank based method is adopted for exhibit the overall performance of the proposed feature level fusion. Matching is performed between a probe fused vector with itself encoded in the database and also with the rest of the encoded fused vectors in the database. The proposed multimodal system is able to identify the specific person from the entire database and ranks are found in terms matching probability obtained. The subjects are retrieved from database according to matching scores. The identification rate for the proposed system is obtained as 98.71% while that for fingerprint and ear biometrics are found to be 95.02% and 93.63% respectively, as shown in Figure 4.

## 7  Conclusion

This paper has presented a feature level fusion technique of fingerprint and ear biometrics for human identification. The technique has used SIFT descriptor for invariant features extraction from fingerprint and ear modalities and PAM characterized K-medoids algorithm for feature reduction. The reduced feature set reflects higher matching proximity with relevant information. Doddington's user-dependent weighting scheme has been adopted by extending the existing notions using adaptive weighting applied to the matching scores. The performance of the technique has been determined on a multimodal database containing fingerprint and ear images. The results show significant improvements on identification performance over the fingerprint and ear biometrics as well as the existing feature level fusion scheme [2] which have used SIFT as feature descriptor. The technique not only attains higher accuracy, but also reflects robustness towards identification of individuals.